\documentclass{article} 
\usepackage{colm2024_conference}
\usepackage{booktabs}
\usepackage{graphicx}
\usepackage{enumitem}
\usepackage{wrapfig}
\usepackage{algorithm}
\usepackage{algpseudocode}
\usepackage{natbib}
\usepackage{makecell}
\usepackage{booktabs}
\usepackage{array}
\usepackage{amsmath}
\usepackage{amssymb}
\usepackage{amsfonts}
\usepackage{multirow}
\usepackage{verbatim}
\usepackage{caption}
\usepackage{longtable}
\usepackage{supertabular}
\usepackage{CJKutf8}
\usepackage[utf8]{inputenc} 
\usepackage[T1]{fontenc}
\usepackage[french,vietnamese,mongolian,greek,english]{babel}
\usepackage{pifont}
\usepackage{tabularx}
\usepackage{geometry}
\usepackage{enumitem}
\usepackage{tablefootnote}
\usepackage{xspace}
\usepackage{textcomp}
\usepackage{makecell}
\usepackage{lscape} 
\usepackage{siunitx}
\usepackage{listings}
\usepackage{xcolor}
\usepackage{svg}
\definecolor{dt}{gray}{0.7}
\usepackage{pifont}       
\usepackage{bbding}       
\usepackage{fontawesome}

\usepackage{scrextend}

\usepackage{tgpagella}
\usepackage{latexsym}
\usepackage[T1]{fontenc}
\usepackage[utf8]{inputenc}
\usepackage{microtype}
\definecolor{mydarkblue}{rgb}{0,0.08,0.45}
\definecolor{citecolor}{HTML}{0071BC}
\usepackage{url}            
\usepackage{nicefrac}       
\usepackage{changepage}
\usepackage{xargs}          
\usepackage{wrapfig,lipsum,booktabs}
\usepackage{longtable}
\usepackage{subcaption}
\usepackage{endnotes}

\usepackage{pgfplots}
\usetikzlibrary{pgfplots.groupplots}
\pgfplotsset{compat=1.3}
\usepackage{tikz}
\usetikzlibrary{patterns}

\usepackage[most]{tcolorbox}

\usepackage{hyperref}
\definecolor{darkblue}{rgb}{0, 0, 0.5}
\hypersetup{colorlinks=true, citecolor=darkblue, linkcolor=darkblue, urlcolor=darkblue, bookmarks=true, bookmarksopen=false, bookmarksnumbered=true}

\usepackage[capitalize,noabbrev]{cleveref}

\crefname{section}{\S}{\S\S}
\Crefname{section}{\S}{\S\S}
\crefname{subsection}{\S\S}{\S\S}
\Crefname{subsection}{\S\S}{\S\S}
\crefformat{section}{\S#2#1#3}
\crefformat{subsection}{\S\S#2#1#3}

\crefname{table}{Table}{Tables}
\crefname{figure}{Figure}{Figures}
\crefname{algorithm}{Algorithm}{}
\crefname{equation}{eq.}{}
\crefname{appendix}{Appendix}{}
\crefformat{section}{Section #2#1#3}
\usepackage{multicol}
\usepackage{tcolorbox}
\usepackage{titlesec}
\titleformat*{\section}{\large\bfseries}
\usepackage{array} 
\newcolumntype{P}[1]{>{\centering\arraybackslash}p{#1}} 
\usepackage{multirow}

\usepackage{adjustbox}
\definecolor{objblue}{RGB}{3,139,221}  
\definecolor{attrred}{RGB}{255,67,67}    
\definecolor{easygreen}{RGB}{0,156,75}  
\definecolor{middleyellow}{RGB}{242,89,34}  
\definecolor{hardred}{RGB}{216,56,58}
\usepackage{colortbl}
\usepackage{array}

\usepackage[most]{tcolorbox}
\definecolor{BoxBackground}{RGB}{240, 240, 240} 
\definecolor{BoxFrame}{RGB}{0, 0, 0} 
\definecolor{TitleBackground}{RGB}{0, 0, 0} 
\definecolor{TitleText}{RGB}{255, 255, 255} 
\tcbset{
  academicbox/.style={
    boxsep=5pt,
    left=2pt,
    right=2pt,
    bottom=0.5pt,
    boxrule=0.5pt,
    colback=BoxBackground,
    colframe=BoxFrame,
    colbacktitle=TitleBackground,
    coltitle=TitleText,
    enhanced,
    attach boxed title to top left={yshift=-0.1in,xshift=0.1in},
    boxed title style={boxrule=0pt,colframe=white},
    title={#1},
  }
}
\newtcolorbox{AcademicBox}[1][]{academicbox=#1}

\title{Qwen-Image-VAE-2.0 Technical Report}

\author{
\bf Qwen Team}

\begin{document}

\maketitle
\vspace{-3mm}

\begin{abstract}
We present Qwen-Image-VAE-2.0, a suite of high-compression Variational Autoencoders (VAEs) that achieve significant advances in both reconstruction fidelity and diffusability\footnote{Diffusability describes how easily a distribution can be modeled by diffusion.}. 
To address the reconstruction bottlenecks of high compression, we adopt an improved architecture featuring Global Skip Connections (GSC) and expanded latent channels. Moreover, we scale training to billions of images and incorporate a synthetic rendering engine to improve performance in text-rich scenarios. 
To tackle the convergence challenges of high-dimensional latent space, we implement an enhanced semantic alignment strategy to make the latent space highly amenable to diffusion modeling.
To optimize computational efficiency, we leverage an asymmetric and attention-free encoder-decoder backbone to minimize encoding overhead.
We present a comprehensive evaluation of Qwen-Image-VAE-2.0 on public reconstruction benchmarks. To evaluate performance in text-rich scenarios, we propose OmniDoc-TokenBench, a new benchmark comprising a diverse collection of real-world documents coupled with specialized OCR-based evaluation metrics. Qwen-Image-VAE-2.0 achieves state-of-the-art reconstruction performance, demonstrating exceptional capabilities in both general domains and text-rich scenarios at high compression ratio.
Furthermore, downstream DiT experiments reveal our models possess superior diffusability, significantly accelerating convergence compared to existing high-compression baselines. 
These establish Qwen-Image-VAE-2.0 as a leading model with high compression, superior reconstruction, and exceptional diffusability.
\end{abstract}


\section{Introduction}
Latent Diffusion Models (LDMs) have become the dominant paradigm in image synthesis~\citep{rombach2022high, flux2024, qwenimage, sd3, seedream3, qwen-image-2.0}. These models typically employ a Variational Autoencoder (VAE) to project images into a compressed latent space for efficient diffusion modeling, where a widely adopted spatial compression ratio is 8 (denoted as $f8$). However, as the industry shifts toward native high-resolution synthesis, this standard ratio faces significant computational bottlenecks. Increasing the spatial compression ratio has thus become essential for computational efficiency, as the complexity of modern Diffusion Transformers (DiTs)~\citep{dit} scales quadratically with the number of latent tokens. Over the past few years, several advances have been achieved in this field, demonstrating the significant potential of high-compression VAEs~\citep{ltx, dcae, cosmos, dcae1.5}.

Despite these advances, a critical challenge exists: the inevitable trade-off between high compression ratio, reconstruction fidelity, and diffusability~\citep{diffusability}. Specifically, higher compression ratios often lead to severe degradation in reconstruction quality, particularly in text-rich scenarios where fine-grained detail is lost. While increasing the latent channel dimension can mitigate this information bottleneck, it frequently results in an over-complex and unstructured latent distribution, which significantly hinders the convergence and generative performance of downstream diffusion models~\citep{vavae, qiu2025image}.

In this work, we introduce Qwen-Image-VAE-2.0, a series of high-compression image VAEs ($f16$ \& $f32$), designed to overcome these challenges through improved architecture, comprehensive data engineering, and enhanced training strategy.

To address the challenge of reconstruction fidelity in high-compression regimes, we adopt an improved VAE architecture with Global Skip Connection (GSC), which establishes a global shortcut from pixels to latents, preserving fine-grained detail. Moreover, our design incorporates a higher latent dimensionality to alleviate the information bottleneck inherent in high-compression scenarios. On the data front, we scale our training corpus to billions of images and curate a specialized document collection (including academic papers, posters, slides, web pages, etc.) to enhance the reconstruction of text-rich images. Furthermore, we develop a synthetic pipeline that renders documents to provide dense supervisory signals for character-level reconstruction. Through these advancements, Qwen-Image-VAE-2.0 achieves state-of-the-art reconstruction performance, especially in text-rich scenarios, despite its high compression ratio.
To address the challenge of diffusability brought by high compression ratio and expanded latent dimension, we demonstrate that semantic alignment with intermediate features of DINOv2~\citep{dinov2} can effectively accelerate DiT convergence. Furthermore, we adopt a staged semantic alignment paradigm that transitions from strict semantic alignment to a balanced optimization of reconstruction and generation. Leveraging these techniques, Qwen-Image-VAE-2.0 achieves superior diffusability compared to existing high-compression VAEs, despite its large channel dimension.

To ensure computational efficiency, we leverage an asymmetric architecture that features a lightweight encoder to minimize encoding overhead during diffusion training. Additionally, we utilize an attention-free backbone to maintain high throughput even with ultra-high-resolution inputs. 

We conduct a comprehensive evaluation to validate the performance of Qwen-Image-VAE-2.0, focusing on two key aspects: reconstruction fidelity and latent space diffusability. For reconstruction fidelity, we evaluate it across general reconstruction tasks and introduce OmniDoc-TokenBench, a benchmark specifically targeting challenging scenarios like real-world text-rich document reconstruction. For latent space diffusability, we also perform extensive downstream DiT experiments to empirically verify it. The results demonstrate that Qwen-Image-VAE-2.0 not only achieves superior reconstruction fidelity, especially in text-rich scenarios despite high compression ratios, but also exhibits excellent latent space compatibility, facilitating rapid DiT convergence even with large latent dimension.

The key contributions of Qwen-Image-VAE-2.0 can be summarized as follows:
\begin{itemize}
\item \textbf{High-Compression VAE}: We introduce a suite of $f16$ and $f32$ image VAEs, providing a robust solution for efficient and native high-resolution image generation.
\item \textbf{Superior Reconstruction Performance:} Qwen-Image-VAE-2.0 achieves state-of-the-art reconstruction performance across multiple benchmarks. It maintains exceptional legibility in text-rich scenarios where traditional high-compression models typically fail.
\item \textbf{Enhanced Latent Diffusability:} Through a refined semantic alignment strategy, we demonstrate that large-channel VAEs can achieve excellent diffusability. This provides a promising solution to the tripartite trade-off between compression ratio, reconstruction fidelity, and diffusability.
\end{itemize}

\section{Model}

In this section, we present the detailed design and architectural innovations of Qwen-Image-VAE-2.0.

\subsection{Design Principle: High Compression VAE with Large Channel}

To optimize the training efficiency of downstream DiTs, we prioritize a higher spatial compression ratio. Given an input image $I \in \mathbb{R}^{H \times W \times 3}$, the VAE maps it to a latent representation $z \in \mathbb{R}^{\frac{H}{f} \times \frac{W}{f} \times C}$, where $f$ denotes the spatial compression ratio and $C$ represents the channel dimension. 
This results in a sequence length of $L = HW/f^2$ for the DiT. Since DiT's computational complexity scales quadratically with sequence length~\citep{attention}, the computation overhead of $\mathcal{O}(L^2)=\mathcal{O}(H^2W^2/{f^4})$ presents a significant bottleneck in high-resolution image generation.

To alleviate this, we move beyond the conventional $f8$ paradigm~\citep{rombach2022high, flux2024, wan2025wan}, adopting higher compression ratios of $f16$ and $f32$ to significantly reduce DiT training costs. While high spatial compression ratio $f$ promises training efficiency, it inevitably constrains the information capacity of the latent space, often resulting in the loss of fine-grained structural detail. To mitigate this, we rely on the principle that reconstruction fidelity is largely governed by the total information bottleneck $N(z)={CHW}/{f^2}$~\citep{dcae}. By increasing the channel dimension $C$, we compensate for the spatial information loss incurred by high compression ratio $f$. Notably, channel expansion does not compromise DiT training efficiency: during training, the DiT first projects latents into a fixed hidden dimension via a linear layer, ensuring that the computational complexity remains nearly invariant to channel dimension.

\subsection{Model Architecture}
\begin{figure}[htbp]
    \centering
    \includegraphics[width=1.0\linewidth]{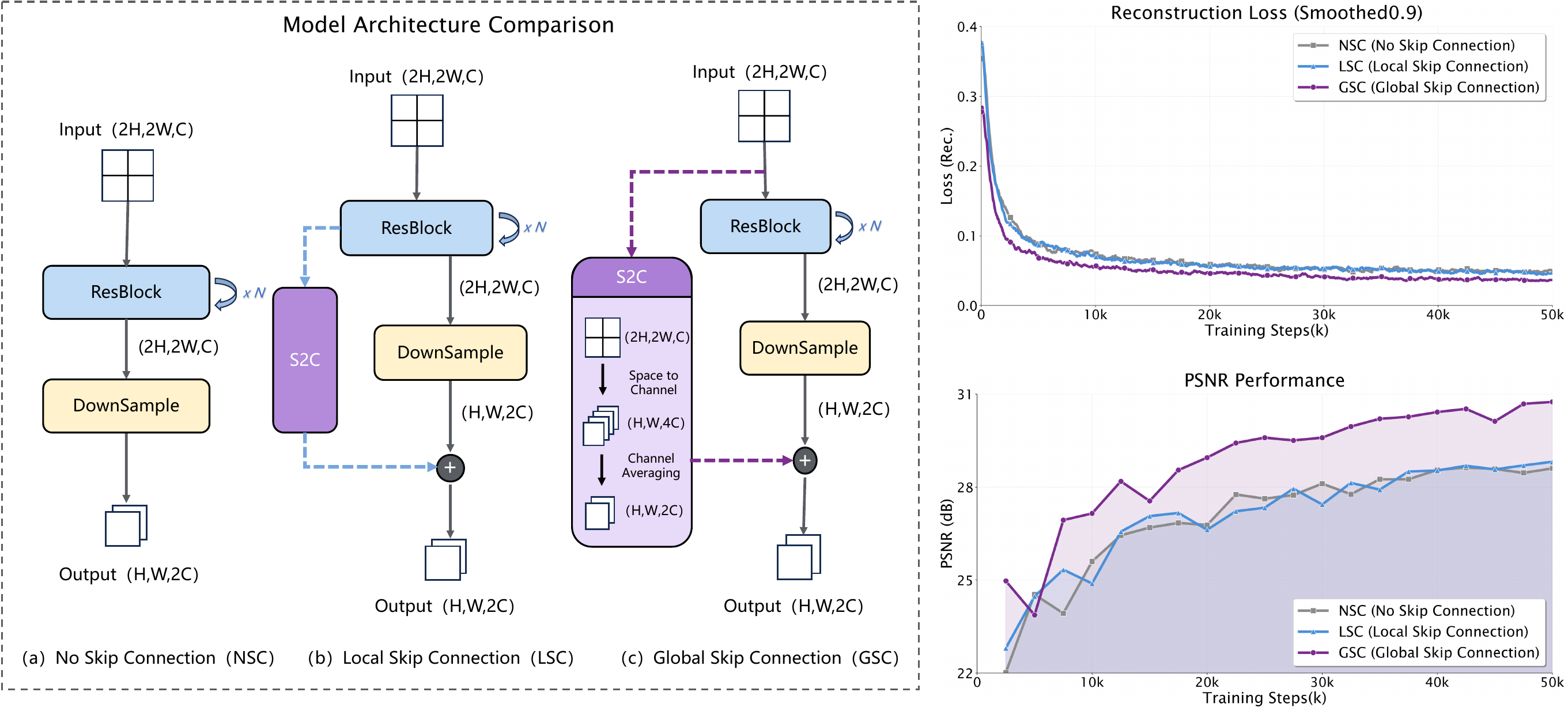}
    \caption{Comparison of No Skip Connection (NSC), Local Skip Connection (LSC), and Global Skip Connection (GSC) on model architecture, reconstruction loss and PSNR performance. S2C is the abbreviation of Space to Channel module. The experiments are conducted on $f16c64$ model training from scratch.}
    \label{fig:vae_arch}
\end{figure}

\paragraph{Global Skip Connection (GSC).}
A primary challenge in high-compression VAEs is the preservation of fine-grained detail during the aggressive downsampling process. The encoder, particularly its non-parametric downsampling layers, often struggles to retain high-frequency information from the original image, leading to optimization difficulties and blurry reconstructions~\citep{dcae}. To alleviate this information loss, we introduce the Global Skip Connection (GSC).

The GSC establishes a direct residual path that bypasses the initial downsampling stage, feeding pixel-level information directly into the deeper latent space. As illustrated in Figure~\ref{fig:vae_arch}, we implement this by employing a space-to-channel operation followed by reshaping, which effectively "folds" spatial information from the input image into the channel dimension.

To validate the efficacy of this design, we conducted an ablation study comparing three configurations: No Skip Connection (NSC), Local Skip Connection (LSC), and Global Skip Connection (GSC). Experiments on an f16c64 model trained from scratch demonstrate that the GSC significantly accelerates convergence by providing the network with high-frequency signal from the input. Based on these findings, we integrate the GSC across the entire Qwen-Image-VAE-2.0 series.

\paragraph{Attention-Free Backbone.} 
For an input of sequence length $N$, the computational complexity of self-attention is $\mathcal{O}(N^2)$, whereas that of convolution with a kernel size $k$ is $\mathcal{O}(N \cdot k^2)$. This quadratic scaling with pixels creates a significant throughput bottleneck for high-resolution image processing. Moreover, the activation memory required for self-attention also scales as $\mathcal{O}(N^2)$, imposing a severe memory constraint during training. In addition, we observed no significant performance degradation when removing attention modules. Consequently, we adopt an attention-free backbone for the entire Qwen-Image-VAE-2.0 series to ensure both training efficiency and scalability.

\paragraph{Encoder-Decoder Asymmetry.} 
We adopt an asymmetric architecture to balance encoding speed with reconstruction quality. By employing a lightweight encoder, we streamline the latent extraction process, effectively reducing training latency for the downstream DiT. Meanwhile, the heavyweight decoder guarantees high-fidelity reconstruction and the preservation of intricate image detail.

\subsection{Model Configurations}

The detailed configurations of the Qwen-Image-VAE-2.0 series are summarized in Table~\ref{tab:vae_configs}.

\begin{table}[htbp]
\centering
\caption{Configurations of Qwen-Image-VAE-2.0 suite. $d_{enc}$ and $d_{dec}$ denote the first projected hidden dimensions of the encoder and decoder. $n_{\text{layer}}$ denotes the number of layers.}
\label{tab:vae_configs}
\begin{tabular}{lccccccc}
\toprule
\textbf{Model} & $f$ & $C$ & $d_{enc}$ & $d_{dec}$ & $n_{\text{layer}}$ & \textbf{Residual} & \textbf{\#Params (Enc/Dec)} \\ \midrule
Qwen-Image-VAE-2.0-f16c64  & 16 & 64  & 96 & 144 & 5 & GSC & 76M / 248M \\
Qwen-Image-VAE-2.0-f16c128  & 16 & 128  & 96 & 144 & 5 & GSC & 76M / 248M \\
Qwen-Image-VAE-2.0-f32c128 & 32 & 128 & 96 & 144 & 6 & GSC & 77M / 250M \\
Qwen-Image-VAE-2.0-f32c192 & 32 & 192 & 96 & 144 & 6 & GSC & 78M / 250M \\ \bottomrule
\end{tabular}
\end{table}

\section{Data}
\subsection{Data Collection}
\paragraph{Scaling Data to Billion Scale.} To ensure robust generalization across diverse domains, we scale the VAE training corpus to encompass billions of images. This large-scale dataset covers a wide spectrum of visual content, spanning various categories, resolutions and aspect ratios. However, data at this scale inevitably contains noise like edge blur and compression artifacts, which impedes model's ability to learn high-frequency detail. To mitigate this, we employ clarity and blur filters to prune low-quality samples, ensuring that the VAE is supervised by high-fidelity signals.

\paragraph{Text-Rich Image Collection.} To address the reconstruction bottleneck in text-rich scenarios, we adopt a two-fold strategy. First, we leverage an OCR filter to identify and prioritize samples with high character density from extensive real-world datasets. Second, we curate a specialized document corpus, which includes screenshots of academic papers, presentation slides, posters, and complex web pages. By training on these real-world text-rich images, our models learn to prioritize the preservation of sharp edges of characters and semantic structures, enabling legible text reconstruction that is challenging for high compression VAEs.

\subsection{Data Synthesis}
To further enhance character-level reconstruction, we develop a synthetic pipeline that renders text documents into images. Our pipeline supports both alphabetic (English) and logographic (Chinese) languages, accounting for their distinct stroke densities and complexities. Crucially, we observed models trained on background-free synthetic data (e.g., black text on white backgrounds) generalize poorly to real-world images where text is often overlaid on complex textures. To bridge this gap, we implement background-contained synthesis, where text is rendered onto backgrounds randomly sampled from general-domain images. Moreover, to adapt different compression settings, we construct synthetic datasets of varying difficulty by rendering characters ranging from 5 to 20 pixels. This multi-granularity supervision forces the VAE to capture fine detail, ensuring legibility even at $f32$ compression.

\section{Training}

\subsection{Training Loss}
The training objective of our image VAE is designed to be simple yet effective. Unlike traditional VAE frameworks that introduce distributional priors and adversarial paradigms, our training process focuses on high-fidelity reconstruction and semantic alignment of the latent space.

The total training loss $\mathcal{L}_{total}$ is formulated as:
\begin{equation}
\mathcal{L}_{total} = \mathcal{L}_{recon} + \lambda_{lpips} \mathcal{L}_{lpips} + \lambda_{align} \mathcal{L}_{align},
\end{equation}
where $\mathcal{L}_{recon}$ is the pixel-level $L_1$ reconstruction loss, and $\mathcal{L}_{lpips}$ denotes the perceptual loss~\citep{lpips} weighted by $\lambda_{lpips}$. To improve the ``diffusability'' of the latent space, we incorporate a semantic alignment loss $\mathcal{L}_{align}$ which aligns latents to semantic counterparts extracted from pretrained encoders (detailed in Sec~\ref{sec:align}). This ensures that the latent space captures more generation-friendly features.

Our empirical findings suggest that two common practices in VAE training (KL regularization and adversarial training) can be removed to achieve better performance and training stability.

\paragraph{Removing KL Loss for Enhanced Semantic Alignment.} 
We remove the Kullback-Leibler (KL) divergence loss as it inherently restricts latent capacity and compromises reconstruction fidelity. More importantly, we observe that the KL penalty acts as a competing constraint to our semantic alignment objective. Given that target semantic features are not necessarily Gaussian-distributed, forcing the model to satisfy both a normal prior and a semantic manifold leads to suboptimal alignment, which ultimately delays the convergence of the downstream DiT. By removing this constraint, we achieve a more flexible latent space that is better suited for generative tasks.

\paragraph{Removing GAN Loss for Training Stability and Efficiency.} 
While GAN loss~\citep{gan} are conventionally used to sharpen visual detail, we find them unnecessary when the training budget is sufficiently large. Given extensive data and iterations, the combination of $\mathcal{L}_{recon}$ and $\mathcal{L}_{lpips}$ is capable of producing high-quality and sharp reconstructions. Eliminating the discriminator not only simplifies the optimization landscape, but also improves training stability and accelerates the overall training process.

In summary, by breaking the conventional reliance on KL loss and GAN loss, we demonstrate the feasibility and effectiveness of a simplified training objective, providing insights for future VAEs.

\subsection{Semantic Alignment}\label{sec:align}
Inspired by~\citet{vavae}, we introduce a semantic alignment loss to strike a delicate balance between low-level detail preservation and high-level semantics, thereby making the latent space more generation-friendly.

\paragraph{Selection of Semantic Encoder.}
Through extensive ablation studies comparing various pretrained vision encoders (including DINOv2~\citep{dinov2}, DINOv3~\citep{dinov3}, MAE~\citep{mae}, and PE-Spatial~\citep{bolya2025perception}), we find that DINOv2 consistently outperforms other candidates in providing generation-friendly semantic priors. Consequently, we select DINOv2-L features as our default semantic guidance.

\paragraph{Selection of Aligned Layer.}
We observe that the choice of encoder layer affects the alignment results. While conventional approaches often utilize the final layer, we find that middle layer of these encoders offer smoother spatial maps that are easier to align with, yielding more generation-friendly latent space. Furthermore, we find that naively combining features from different layers introduces unnecessary noise that corrupts the alignment signal. Consequently, we align the VAE latent with a single, optimally selected middle layer, rather than relying on the final output or multi-layer fusion.

Specifically, given a target image, we first extract the semantic feature map $f \in \mathbb{R}^{h \times w \times c}$ using the pretrained semantic encoder, where $h$ and $w$ denote the spatial resolution and $c$ is the feature dimension. Then, we project the VAE latent $z$ into the same dimensionality through a learnable linear transformation, $z' = Wz$, where $W$ is a trainable projection matrix. Let $\mathcal{P} = \{(i,j) \mid 1 \le i \le h,\; 1 \le j \le w\}$ denote the set of spatial positions, where $|\mathcal{P}| = N = hw$. For each position $p \in \mathcal{P}$, we denote by $f_p \in \mathbb{R}^{c}$ and $z'_p \in \mathbb{R}^{c}$ the semantic feature and projected latent feature at position $p$, respectively.

The alignment objective consists of two complementary components: (1) a Marginal Cosine Similarity Loss $\mathcal{L}_{mcos}$ with margin $m_{cos}$ which aligns the direction of VAE latents with target semantics, and (2) a Marginal Distance Matrix Similarity Loss $\mathcal{L}_{mdms}$ with margin $m_{dist}$, which preserves the relative spatial layout. The core alignment objectives are formulated as:

\begin{align}
\mathcal{L}_{mcos}(z', f)
&= \frac{1}{N} \sum_{p \in \mathcal{P}} \mathrm{ReLU} \left( 1 - \cos(z'_p, f_p) - m_{cos}\right), \\
\mathcal{L}_{mdms}(z', f)
&= \frac{1}{N^2} \sum_{p \in \mathcal{P}} \sum_{q \in \mathcal{P}} \mathrm{ReLU} \left( \left| \cos(z'_p, z'_q) - \cos(f_p, f_q) \right| - m_{dist} \right), \\
\mathcal{L}_{align}(z, f)
&= \mathcal{L}_{mcos}(z', f) + \mathcal{L}_{mdms}(z', f),
\end{align}

where $\mathcal{P}$ denotes the set of spatial positions and $N=hw$ is the total number of elements.

\subsection{Training Strategy}
We employ a multi-stage training paradigm designed to progressively improve spatial resolution, refine textual rendering, and ensure robust semantic alignment.

\paragraph{Enhancing Resolution: From Low Resolution to High Resolution.} 
To facilitate stable training, we adopt a curriculum-based resolution strategy, starting from low-resolution foundations and progressively scaling up to 2K. Throughout this progression, we incorporate a diverse spectrum of aspect ratios to enhance the model's geometric robustness, ensuring the VAE maintains structural integrity across various image compositions without distortion. This progressive upscaling allows the model to first learn basic structures and then capture finer detail and textures.

\paragraph{Integrating Textual Rendering: From Non-text to Text.} 
To master high-fidelity text reconstruction, we employ a multi-stage data infusion strategy. We start with general-domain images to accelerate initial convergence. Subsequently, we progressively incorporate real-world text-rich samples to address the challenges of complex character recognition. In the final phase, we introduce synthetic text data across varying difficulty levels to refine character precision. As general textures and character detail require different reconstruction focuses, we maintain a balanced ratio between these two types of data to ensure high quality for both.

\paragraph{Calibrating Semantic Alignment: From Strict Alignment to Balanced Reconstruction.} 
At the beginning of training, we apply strict semantic alignment using a strict margin ($m_{cos}$ and $m_{dist}$). We found that strong alignment at the early stage significantly helps the diffusability of the latent space. As the training progresses, we gradually loose the alignment margins. This allows the model to strike a better balance between maintaining semantic consistency and achieving high-quality pixel-level reconstruction.

\section{OmniDoc-TokenBench}
\label{sec:omnidoc_tokenbench}

\subsection{Motivation}
Standard reconstruction benchmarks such as ImageNet~\citep{deng2009imagenet} and FFHQ~\citep{Karras2018ASG} consist predominantly of natural photographs with negligible textual content, making them ill-suited for evaluating text-rich image reconstruction. Conventional pixel-level metrics (PSNR, SSIM) are inherently insensitive to text legibility, as minor stroke distortions may lead to a negligible decrease in conventional evaluation metrics yet render characters unrecognizable. While TokBench~\citep{Wu2025TokBenchEY} introduces OCR-based reconstruction evaluation, its data is drawn from scene text datasets where text instances are sparse and character sizes are insufficiently small, making it inadequate for benchmarking reconstruction capability in text-rich scenarios.

\begin{figure}[htbp]
    \centering
    \includegraphics[width=\linewidth]{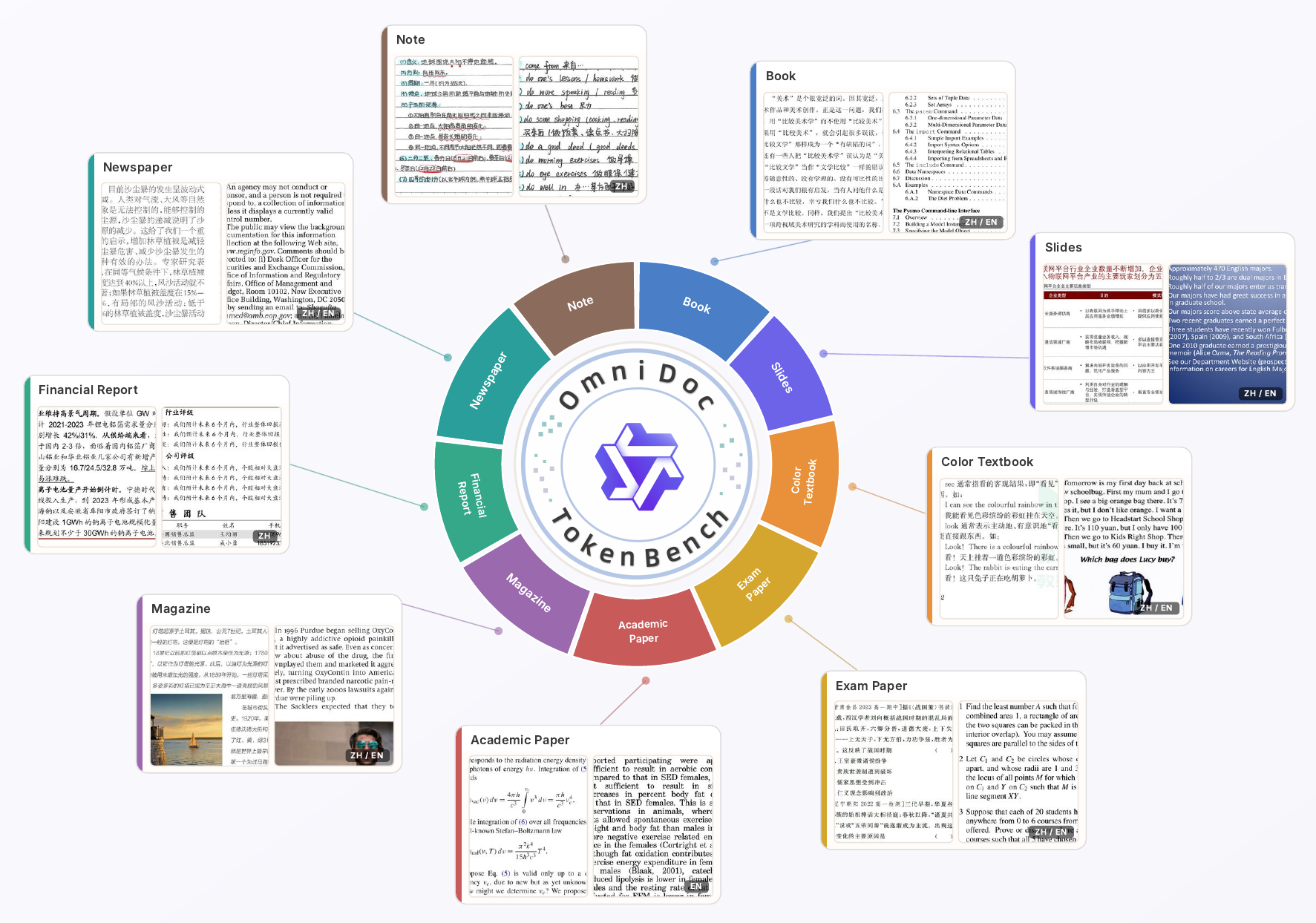}
    \caption{OmniDoc-TokenBench, a curated collection of ${\sim}$3K text-rich images. }
    \label{fig:vae_bench}
\end{figure}

To address these limitations, we propose \textbf{OmniDoc-TokenBench} (Figure~\ref{fig:vae_bench}), a curated benchmark of ${\sim}$3K text-rich document images spanning nine categories---\textit{book}, \textit{slides}, \textit{color textbook}, \textit{exam paper}, \textit{academic paper}, \textit{magazine}, \textit{financial report}, \textit{newspaper}, and \textit{note}---covering both alphabetic (English) and logographic (Chinese) text. We perform full-page OCR on both the original and reconstructed images and compute Normalized Edit Distance (\textbf{NED})~\citep{Liu2019ICDAR2R, Marzal1993ComputationON} between their OCR outputs, directly measuring page-level document readability without requiring word-level bounding box annotations. This annotation-free design also facilitates easy scaling to new document types.

\subsection{Benchmark Construction}
OmniDoc-TokenBench is derived from OmniDocBench~\citep{Ouyang2024OmniDocBenchBD}, a document parsing dataset offering fine-grained layout annotations and text-level ground truth across diverse sources. We construct the benchmark through a four-stage pipeline:

\paragraph{Text block extraction and font normalization.}
Specifically, we first crop a region from the top-left corner of each text block and then resize it to $256\!\times\!256$ pixels so that each character occupies approximately 
$f_{\text{ref}}\!\times\!f_{\text{ref}}$ pixels. We set 
$f_{\text{ref}}\!=\!16$ for Chinese and  $f_{\text{ref}}\!=\!10$ for English. These reference sizes are chosen empirically to provide a meaningful evaluation regime: the resulting inputs remain challenging for VAE reconstruction, especially in preserving fine stroke details, while standard OCR models still maintain high recognition accuracy.

\paragraph{Content filtering.}
We apply PP-OCRv5~\citep{cui2025paddleocr30technicalreport} to each sample, and retain only samples whose total count of recognized characters fall within $[200, 600]$ (Chinese) or $[300, 600]$ (English), ensuring sufficient textual density for reliable metric computation while excluding overly sparse or dense samples.

\paragraph{Deduplication.}
We compute character-level $n$-gram overlap ($n\!=\!3$ for Chinese, $n\!=\!5$ for English) between samples from the same source page and across the same document category. Pairs exceeding overlap thresholds of $0.2$ (intra-page) or $0.3$ (intra-category) are considered duplicates, among each overlapping group, only the sample with the highest character count is retained.

\paragraph{Human inspection.}
To ensure data quality, we manually prune samples that are blurred, visually redundant, or contain significant blank regions. The final benchmark maintains a roughly balanced distribution between Chinese and English text.

\subsection{Evaluation Methodology}
Beyond standard reconstruction metrics (PSNR~\citep{psnr}, SSIM~\citep{ssim}, LPIPS~\citep{lpips}, FID~\citep{fid}), we employ \textbf{NED} as the primary text-fidelity metric. A key design choice is using the OCR output of the \emph{original} image as reference rather than the ground-truth annotations. Since OCR models introduce systematic errors even on clean inputs (e.g., confusing visually similar characters such as ``rn'' vs.\ ``m''), comparing against annotations would falsely attribute such errors to the VAEs. By applying the same OCR model to both images, these biases largely cancel in the edit distance computation, isolating degradation caused solely by reconstruction.

Concretely, for each image $I_i$ in the benchmark, we apply PP-OCRv5 independently to the original image and its VAE reconstruction, yielding text strings $s_{\mathrm{gt}}^{(i)}$ and $s_{\mathrm{recon}}^{(i)}$, respectively. The benchmark-level NED is defined as:

\begin{equation}
\mathrm{NED} = \frac{1}{N}\sum_{i=1}^{N}\left(1 - \frac{d_{\mathrm{edit}}\!\bigl(s_{\mathrm{gt}}^{(i)},\, s_{\mathrm{recon}}^{(i)}\bigr)}{\max\!\bigl(|s_{\mathrm{gt}}^{(i)}|,\, |s_{\mathrm{recon}}^{(i)}|\bigr)}\right),
\label{eq:ocr_ned}
\end{equation}

where $d_{\mathrm{edit}}(\cdot,\cdot)$ denotes the Levenshtein distance, $N$ is the total number of benchmark images, and $|\cdot|$ denotes the string length. Each term measures character-level agreement for a single image, averaging over the benchmark yields a robust global estimate of text reconstruction fidelity.

\section{Experiments}

\begin{table}[t]
\centering
\small
\caption{Comparison of different baselines against our Qwen-Image-VAE-2.0 models across various compression settings. Our models are highlighted in \colorbox{blue!5}{purple}. Underline means second best score.}
\label{tab:main_bench}
\resizebox{\textwidth}{!}{%
\begin{tabular}{llccccccc}
\toprule
\multirow{2}{*}{\textbf{Baseline}} & \multirow{2}{*}{\textbf{Setting}} & \multicolumn{2}{c}{\textbf{Generation(w/o CFG)}} & \multicolumn{2}{c}{\textbf{Recon@Imagenet}} & \multicolumn{2}{c}{\textbf{Recon@FFHQ}} \\
\cmidrule(lr){3-4} \cmidrule(lr){5-6} \cmidrule(lr){7-8}
& & IS $\uparrow$ & gFID $\downarrow$ & PSNR $\uparrow$ & SSIM $\uparrow$ & PSNR $\uparrow$ & SSIM $\uparrow$ \\
\midrule
\rowcolor{gray!8} \multicolumn{8}{l}{\textit{ViT-backone AutoEncoders}} \\
VTP-Large~\citep{vtp} & f16c64 & 146.22 & 5.25 & 26.88 & 0.7718 & 16.52 & 0.3129 \\
RAE-DINOv2-B~\citep{rae} & f16c768 & 139.80 & 6.64 & 19.24 & 0.5025 & -- & -- \\
RAE-SigLIP2-B~\citep{rae} & f16c768 & 103.24 & 11.58 & 19.71 & 0.5279 & -- & -- \\
\midrule
\rowcolor{gray!8} \multicolumn{8}{l}{\textit{f8 Compression VAEs}} \\
FLUX.1-dev~\citep{flux2024}       & f8c16    & 54.64  & 25.41 & 32.84 & 0.9155 & 38.14 & 0.9574 \\
HunyuanVideo~\citep{hunyuanvideo}     & f8c16    & 63.57  & 21.29 & 33.21 & 0.9143 & 39.85 & 0.9607 \\
Qwen-Image~\citep{qwenimage}       & f8c16    & 73.52  & 17.68 & 33.42 & 0.9159 & 38.75 & 0.9512 \\
Wan2.1~\citep{wan2025wan}           & f8c16    & 78.60  & 16.25 & 31.29 & 0.8870 & 38.16 & 0.9456 \\
Cosmos-0.1-CI8x8~\citep{cosmos}   & f8c16    & 52.89  & 26.02 & 32.33 & 0.9024 & 39.16 & 0.9546 \\
\midrule
\rowcolor{gray!8} \multicolumn{8}{l}{\textit{f16 Compression VAEs}} \\
Cosmos-0.1-CI16x16~\citep{cosmos}   & f16c16 & 85.14  & 15.21 & 25.13 & 0.7015 & 30.91 & 0.8285 \\
HunyuanVideo-1.5~\citep{hunyuanvideo1.5} & f16c32   & 69.59  & 19.08 & 31.18 & 0.8710 & 37.30 & 0.9336 \\
HunyuanImage-3.0~\citep{hunyuanimage3.0} & f16c32   & 73.84  & 17.87 & 31.08 & 0.8655 & 36.85 & 0.9260 \\
VAVAE~\citep{vavae} & f16c32 & \textbf{129.80} & \textbf{6.03} & 27.75 & 0.7986 & 32.84 & 0.8752 \\
Wan2.2~\citep{wan2025wan}           & f16c48   & 79.55  & 15.65 & 31.30 & 0.8784 & 37.75 & 0.9386 \\
Stepvideo-T2V~\citep{stepvideo}    & f16c64   & 45.18  & 33.53 & 31.54 & 0.8973 & 37.46 & 0.9451 \\
FLUX.2-dev~\citep{flux2}       & f16c128  & 91.53  & 10.61 & 34.34 & 0.9358 & 40.36 & 0.9676 \\
\rowcolor{blue!5} \textbf{Qwen-Image-VAE-2.0-f16c64} & f16c64 & \underline{102.76} & \underline{9.52} & 32.72 & 0.9086 & 39.14 & 0.9541 \\
\rowcolor{blue!5} \textbf{Qwen-Image-VAE-2.0-f16c128} & f16c128 & 92.42 & 10.29 & \textbf{35.90} & \textbf{0.9519} & \textbf{43.10} & \textbf{0.9795} \\
\midrule
\rowcolor{gray!8} \multicolumn{8}{l}{\textit{f32 Compression VAEs}} \\
DC-AE-sana~\citep{dcae}        & f32c32   & \underline{75.73}  & \underline{16.88} & 24.82 & 0.6897 & 31.35 & 0.8303 \\
HunyuanImage-2.1~\citep{hunyuanimage2.1} & f32c64   & 47.96  & 33.32 & 28.67 & 0.8199 & 35.30 & 0.9110 \\
LTX-Video~\citep{ltx}            & f32c128  & 33.48  & 44.94 & 29.57 & 0.8329 & 35.56 & 0.9051 \\
LTX-2~\citep{ltx2}            & f32c128  & 42.57  & 38.19 & 26.06 & 0.7925 & 33.63 & 0.9058 \\
\rowcolor{blue!5} \textbf{Qwen-Image-VAE-2.0-f32c128} & f32c128 & \textbf{81.23} & \textbf{15.05} & \underline{29.69} & \underline{0.8423} & \underline{35.91} & \underline{0.9177} \\
\rowcolor{blue!5} \textbf{Qwen-Image-VAE-2.0-f32c192} & f32c192 & 72.31 & 18.33 & \textbf{31.13} & \textbf{0.8785} & \textbf{37.52} & \textbf{0.9381} \\
\bottomrule
\end{tabular}
}
\end{table}

\subsection{Quantitative Results}
\subsubsection{Performance of Reconstruction}
We evaluate the reconstruction performance of Qwen-Image-VAE-2.0 on two standard benchmarks: ImageNet~\citep{deng2009imagenet} and FFHQ\citep{Karras2018ASG}. Specifically, we evaluate low-resolution (256p) general-domain performance on ImageNet and high-resolution (1K) performance on FFHQ, using the Peak Signal-to-Noise Ratio (PSNR) and the Structural Similarity Index Measure (SSIM) as our quality metrics. As demonstrated in Table~\ref{tab:main_bench}, Qwen-Image-VAE-2.0 achieves state-of-the-art reconstruction fidelity within its corresponding compression tiers ($f16$ and $f32$). Notably, our $f32c192$ VAE performs comparably to established $f8$ VAEs (e.g., Wan2.1), despite operating at a $4\times$ compression factor. This superior performance in reconstruction is largely attributable to our refined VAE architecture, expanded channel dimensions, and extensive training regimen.

\begin{table}[t]
\centering
\tiny
\caption{Comprehensive evaluation on OmniDoc-TokenBench (${\sim}$3K text-rich images, $256\!\times\!256$). Models are grouped by spatial compression factor and sorted by NED within each group. Our models are highlighted in \colorbox{blue!5}{purple}. }
\label{tab:text_bench}
\resizebox{\textwidth}{!}{%
\begin{tabular}{llccccc}
\toprule
\textbf{Model} & \textbf{Setting} & \textbf{SSIM}\,$\uparrow$ & \textbf{PSNR}\,$\uparrow$ & \textbf{LPIPS}\,$\downarrow$ & \textbf{FID}\,$\downarrow$ & \textbf{NED}\,$\uparrow$ \\
\midrule
\rowcolor{gray!8} \multicolumn{7}{l}{\textit{ViT-backbone AutoEncoders}} \\
RAE-DINOv2-B~\citep{rae}            & f16c768   & 0.3261 & 14.32 & 0.2290 & 18.21 & 0.0392 \\
RAE-SigLIP2-B~\citep{rae}           & f16c768   & 0.3871 & 14.36 & 0.1972 & 11.49 & 0.0483 \\
VTP-Large~\citep{vtp}               & f16c64   & \textbf{0.7185} & \textbf{18.11} & \textbf{0.1046} & \textbf{3.94} & \textbf{0.4170} \\
\midrule
\rowcolor{gray!8} \multicolumn{7}{l}{\textit{f8 Compression VAEs}} \\
Wan2.1~\citep{wan2025wan}                  & f8c16     & 0.8282 & 20.54 & 0.0679 & 4.57  & 0.8021 \\
Cosmos-0.1-CI8x8~\citep{cosmos}            & f8c16     & 0.9057 & 24.29 & 0.0464 & 2.89  & 0.9033 \\
Qwen-Image~\citep{qwenimage}               & f8c16     & 0.8998 & 24.94 & 0.0519 & 4.48  & 0.9073 \\
HunyuanVideo~\citep{hunyuanvideo}          & f8c16     & 0.9227 & 25.26 & 0.0434 & 2.03  & 0.9266 \\
FLUX.1-dev~\citep{flux2024}                & f8c16     & \textbf{0.9364} & \textbf{26.24} & \textbf{0.0246} & \textbf{0.55}  & \textbf{0.9546} \\
\midrule
\rowcolor{gray!8} \multicolumn{7}{l}{\textit{f16 Compression VAEs}} \\
Cosmos-0.1-CI16x16~\citep{cosmos}            & f16c16    & 0.5460 & 15.55 & 0.1349 & 7.78  & 0.1547 \\
VAVAE~\citep{vavae}              & f16c32    & 0.6905 & 17.50 & 0.0974 & 4.45  & 0.3488 \\
HunyuanVideo-1.5~\citep{hunyuanvideo1.5}          & f16c32    & 0.8422 & 21.49 & 0.0839 & 4.67  & 0.6938 \\
HunyuanImage-3.0~\citep{hunyuanimage3.0}          & f16c32    & 0.8672 & 22.66 & 0.0650 & 3.49  & 0.7753 \\
Wan2.2~\citep{wan2025wan}                    & f16c48    & 0.8577 & 21.67 & 0.0525 & 3.05  & 0.8310 \\
Stepvideo-T2V~\citep{stepvideo}             & f16c64    & 0.8970 & 23.69 & 0.0650 & 6.02  & 0.8838 \\
\rowcolor{blue!5} \textbf{Qwen-Image-VAE-2.0-f16c64}      & f16c64   & 0.9279 & 26.00 & 0.0382 & 1.94  & 0.9244 \\
FLUX.2-dev~\citep{flux2}                & f16c128   & 0.9544 & 27.72 & 0.0216 & \textbf{0.73}  & 0.9535 \\
\rowcolor{blue!5} \textbf{Qwen-Image-VAE-2.0-f16c128}     & f16c128  & \textbf{0.9706} & \textbf{30.45} & \textbf{0.0167} & 0.79 & \textbf{0.9617} \\
\midrule
\rowcolor{gray!8} \multicolumn{7}{l}{\textit{f32 Compression VAEs}} \\
DC-AE-sana~\citep{dcae}                 & f32c32    & 0.5259 & 15.62 & 0.1441 & 7.26  & 0.0692 \\
LTX-2~\citep{ltx2}                     & f32c128   & 0.7354 & 18.41 & 0.1192 & 9.94  & 0.3569 \\
HunyuanImage-2.1~\citep{hunyuanimage2.1}          & f32c64    & 0.7805 & 19.85 & 0.0957 & 5.19  & 0.4895 \\
LTX-Video~\citep{ltx}                     & f32c128   & 0.8055 & 20.92 & 0.1190 & 17.10 & 0.5651 \\
\rowcolor{blue!5} \textbf{Qwen-Image-VAE-2.0-f32c128}     & f32c128  & 0.8442 & 22.13 & 0.0642 & 3.36  & 0.7065 \\
\rowcolor{blue!5} \textbf{Qwen-Image-VAE-2.0-f32c192}     & f32c192  & \textbf{0.8908} & \textbf{23.84} & \textbf{0.0497} & \textbf{1.98} & \textbf{0.8555} \\
\bottomrule
\end{tabular}
}
\end{table}

\subsubsection{Performance of Text Rendering}
\label{sec:text_rendering}

\begin{figure*}[t]
    \centering
    \begin{subfigure}{\textwidth}
        \centering
        \includegraphics[width=1.0\textwidth]{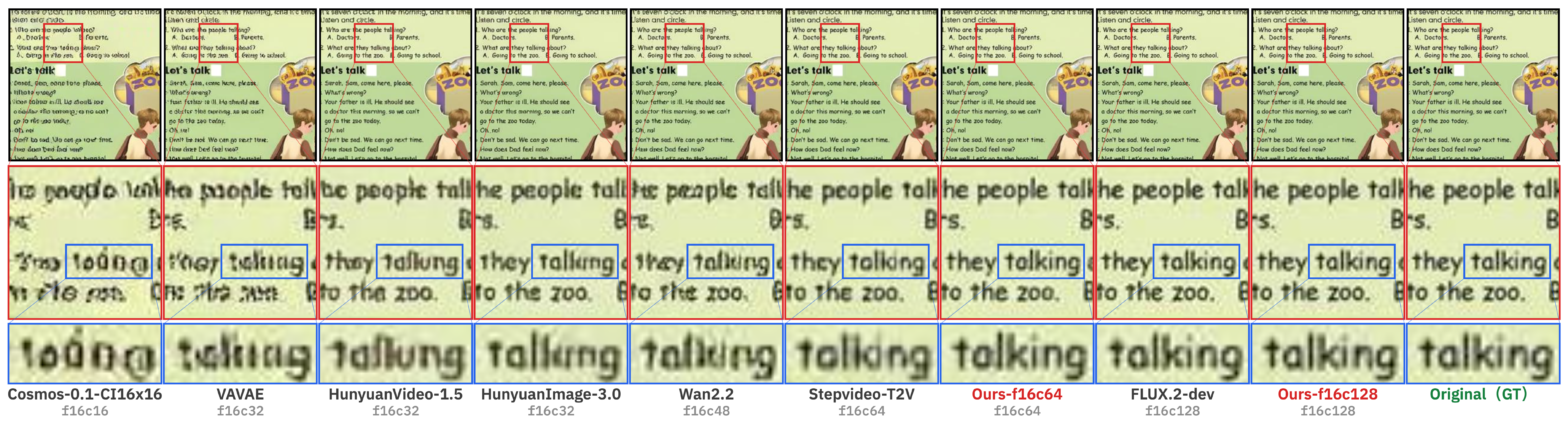}
        \caption{$f16$ Compression VAEs. Baselines exhibit character blurring and stroke merging, while ours \textcolor[RGB]{215,36,36}{$f16$ VAEs} preserves crisp boundaries.}
        \label{fig:f16_comparison}
    \end{subfigure}

    \begin{subfigure}{\textwidth}
        \centering
        \includegraphics[width=1.0\textwidth]{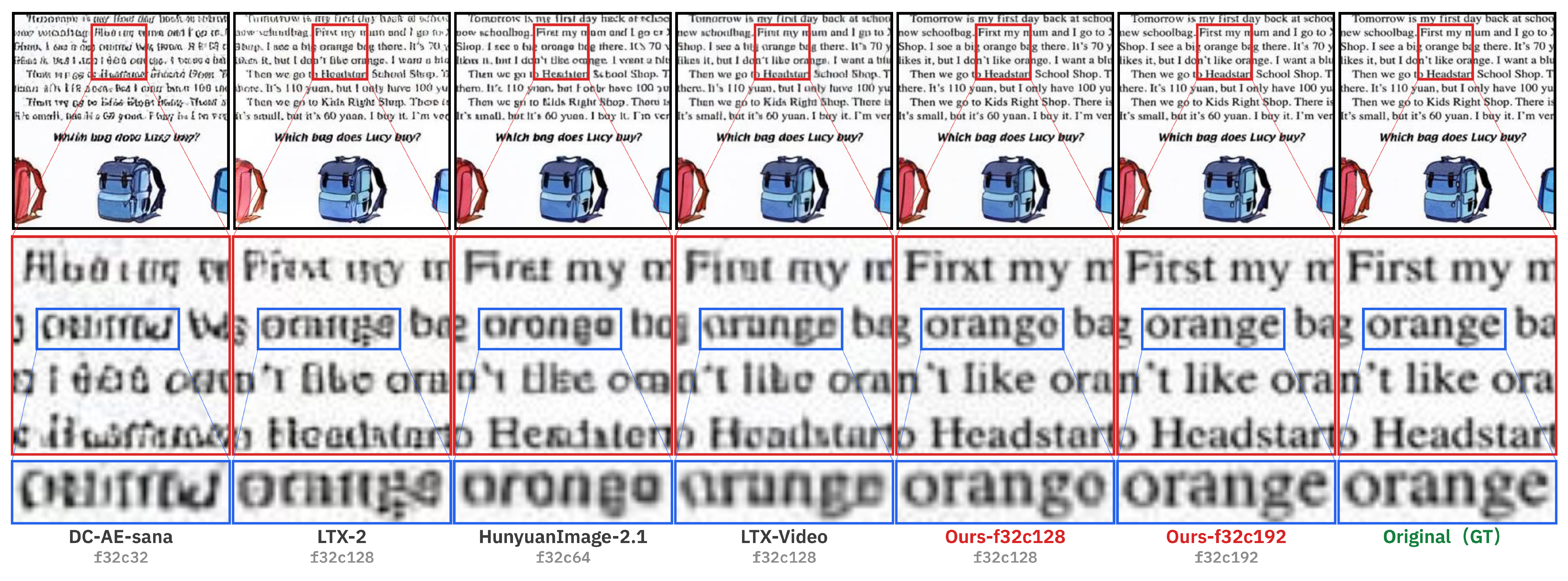}
        \caption{$f32$ Compression VAEs. Competing models reduce text to illegible smears; our \textcolor[RGB]{215,36,36}{$f32$ VAEs} retains distinguishable characters and word boundaries.}
        \label{fig:f32_comparison}
    \end{subfigure}
    
    \caption{Qualitative comparison of text reconstruction on Ours OmniDoc-TokenBench. For each compression ratio, we show: (Top) full image with patch crop regions indicated; (Middle) zoomed-in patch with word box annotations; (Bottom) zoomed-in word.}
    \label{fig:text_recon_comparison}
\end{figure*}

To assess reconstruction fidelity in challenging text-rich scenarios, we evaluate all models on OmniDoc-TokenBench~(\S\ref{sec:omnidoc_tokenbench}), reporting both traditional pixel-based metrics and the proposed OCR-based NED metric. We compute NED on raw OCR outputs without text normalization, while minor spacing artifacts from OCR may inflate edit distance, the evaluation pipeline is applied consistently across all models ensuring fair comparison. Results are summarized in Table~\ref{tab:text_bench}, with qualitative comparisons in Figure~\ref{fig:text_recon_comparison}.

\paragraph{Traditional Reconstruction Metrics.}
Our high-compression VAEs ($f16$ and $f32$ settings) achieve state-of-the-art pixel-level reconstruction, outperforming all competing methods at the same compression ratios. Notably, Qwen-Image-VAE-2.0-f16c128 attains an SSIM of \textbf{0.9706} and PSNR of \textbf{30.45\,dB}, surpassing the best $f8$ baseline (FLUX.1-dev at 0.9364 / 26.24\,dB) despite $2\times$ higher spatial compression. Even our lighter f16c64 variant surpasses most $f8$ baselines (SSIM 0.9279 vs.\ HunyuanVideo's 0.9227), demonstrating strong efficiency at moderate channel budgets. Under extreme $f32$ compression where existing methods degrade catastrophically (SSIM as low as 0.5259), our f32c192 achieves \textbf{0.8908} SSIM with FID \textbf{1.98}---outperforming all $f32$ competitors by substantial margins.

\paragraph{NED for Text Fidelity.}
Traditional pixel metrics are inherently insensitive to character-level legibility. A single-character reconstruction error such as ``orange'' $\to$ ``orango'' incurs negligible PSNR loss ($<$0.5\,dB) yet reduces NED by \textbf{16.7\%}, exposing the semantic corruption that pixel metrics might miss. The NED metric directly measures text preservation by comparing recognized character sequences between original and reconstructed images.

In the $f16$ VAEs, baseline performance varies dramatically: Cosmos-0.1-CI16x16 collapses to NED \textbf{0.1547} (${\sim}$85\% character loss), while others range 0.35--0.88, all below $f8$ state-of-the-art. Our Qwen-Image-VAE-2.0-f16c64 achieves \textbf{0.9244}, competitive with leading $f8$ methods. Most remarkably, Qwen-Image-VAE-2.0-f16c128 reaches \textbf{0.9617}, \emph{surpassing all evaluated $f8$ VAEs} including FLUX.1-dev (0.9546). To the best of our knowledge, this is the first $f16$ autoencoder to achieve text fidelity exceeding $f8$ VAEs.

The $f32$ VAEs further validate our advantage. While competing models exhibit near-total text destruction (NED: 0.07--0.57), our Qwen-Image-VAE-2.0-f32c192 achieves \textbf{0.8555}, surpassing multiple $f16$ baselines. This cross-compression superiority stems directly from our comprehensive data engineering incorporating diverse text-rich documents and synthetic rendering pipelines.

\paragraph{Correlation Analysis.}
We observe an imperfect correlation between pixel metrics and text fidelity. In $f16$, Stepvideo-T2V achieves notably higher NED than HunyuanImage-3.0 (0.8838 vs.\ 0.7753) despite modest SSIM differences (0.8970 vs.\ 0.8672); in $f32$, LTX-Video outperforms HunyuanImage-2.1 in NED (0.5651 vs.\ 0.4895) despite worse FID (17.10 vs.\ 5.19). These discrepancies validate NED as a necessary complementary metric for text reconstruction evaluation.

\subsubsection{Performance of Diffusability}
To empirically validate the diffusability of our learned latent space, we evaluate downstream generative performance by training SiT~\citep{sit} on the ImageNet 256$\times$256 dataset. We strictly adhere to the codebase and default hyperparameter settings of~\citet{repa-e}, reporting generation quality at 80 epochs via the Inception Score (IS)~\citep{inception} and generative FID (gFID). Specifically, we employed the SiT-XL/2 architecture for the $f8$ compression setting, and the SiT-XL/1 architecture for the $f16$ and $f32$ settings. Since higher-dimensional latent space often require larger optimal Classifier-Free Guidance (CFG)~\citep{cfg} scales, and the evaluated VAEs possess varying dimensions, we report only the results without guidance to ensure a fair comparison. As shown in Table~\ref{tab:main_bench}, Qwen-Image-VAE-2.0 demonstrates superior latent space diffusability, consistently outperforming existing high-compression baselines in overall generation quality. Despite their large latent dimensions, our models facilitate rapid DiT convergence. This exceptional diffusability effectively resolves the traditional tripartite trade-off and is primarily driven by our improved semantic alignment strategy and staged alignment paradigm.

\subsection{Qualitative Results}

\subsubsection{Text Rendering}

Figure~\ref{fig:text_recon_comparison} visualizes qualitative reconstruction results across different compression ratios. At $f16$ VAEs (Figure~\ref{fig:f16_comparison}), the degradation gap widens dramatically. Weaker baselines show severe character blurring, stroke merging, and inter-character ghosting---Cosmos-0.1-CI16x16~\citep{cosmos} exhibits near-complete text collapse. In contrast, our Qwen-Image-VAE-2.0-f16c128 preserves crisp character boundaries, accurate inter-character spacing, and fine stroke detail.

\begin{figure}[htbp]
    \centering
    \includegraphics[width=0.90\linewidth]{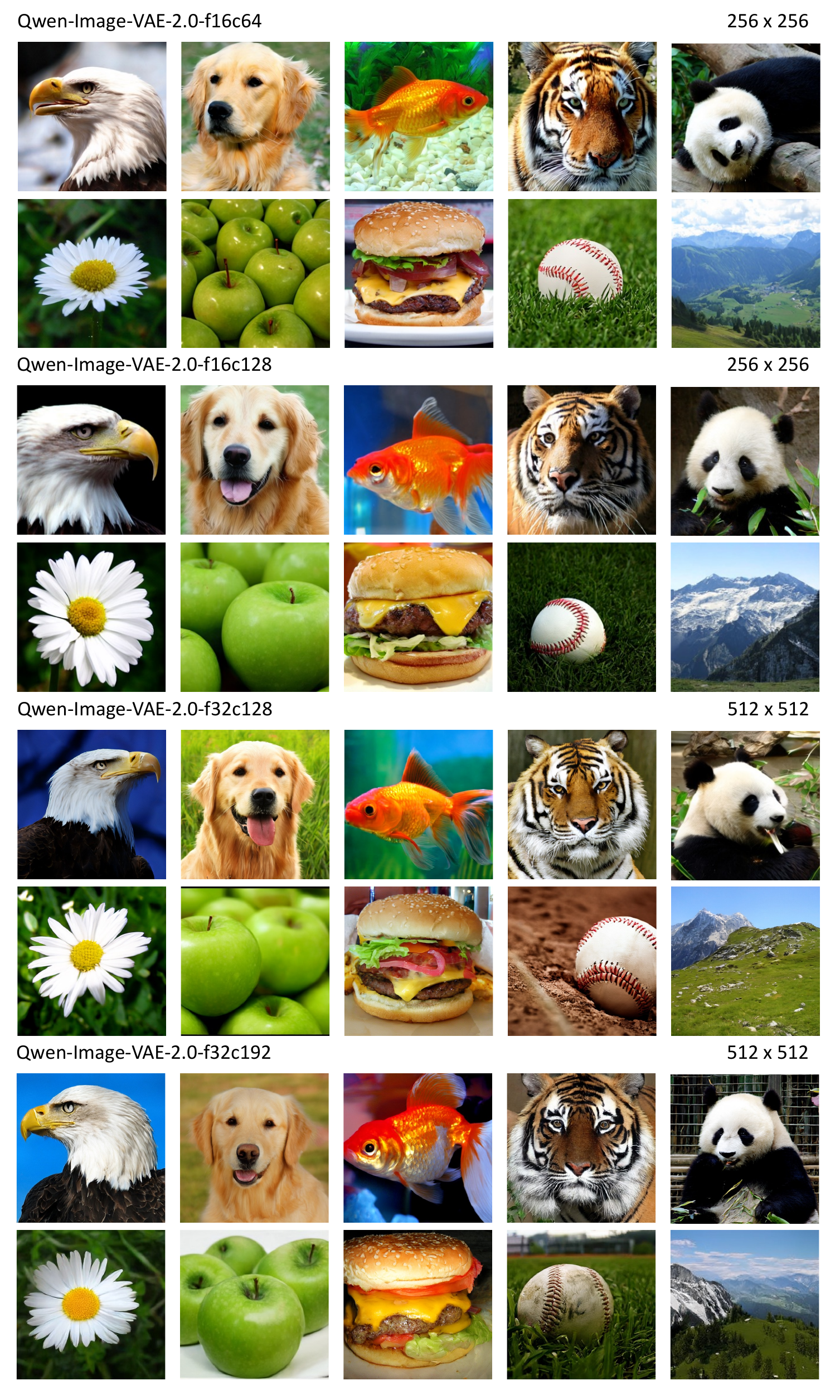}
    \caption{Selected image samples generated by SiT on ImageNet with Qwen-Image-VAE-2.0.}
    \label{fig:sample_images}
\end{figure}

For $f32$ VAEs (Figure~\ref{fig:f32_comparison}), competing models reduce text to fragmented noise patterns where individual characters become unrecognizable. Our Qwen-Image-VAE-2.0-f32c192 uniquely retains clearly distinguishable character forms and recognizable word boundaries, consistent with its substantially higher NED scores and validating the effectiveness of our architecture under extreme compression constraints.

\subsubsection{Diffusability}
Figure~\ref{fig:sample_images} illustrates selected ImageNet samples generated by SiT-XL, serving as a visual validation of the latent space's semantic coherence and fine-grained detail. To demonstrate cross-scale consistency, samples are generated at $256\times256$ for $f16$ VAEs and $512\times512$ for $f32$ VAEs using a further-trained SiT-XL with classifier-free guidance, following our quantitative training protocol. Across $f16$ and $f32$ compression ratios, the generations maintain high visual fidelity without structural degradation.

\subsubsection{Large-scale Text-to-Image Validation}
The successful integration of Qwen-Image-VAE-2.0\footnote{The VAE integrated into Qwen-Image-2.0 is an intermediate variant derived from the methodological framework established in this work.} into the Qwen-Image-2.0~\citep{qwen-image-2.0} further validates the diffusability of our latent space at a foundation-model scale. Operating within this large-scale text-to-image pipeline, our compressed representations readily support complex open-vocabulary conditioning and intricate compositional constraints. The resulting generations consistently demonstrate Qwen-Image-2.0’s core capabilities through precise text rendering and refined photorealistic textures across diverse semantic contexts. This large-scale deployment demonstrates that our latent space preserves the structural stability and information fidelity required for demanding synthesis tasks, thereby confirming the scalability and robustness of our alignment paradigm in advanced generative systems.

\section{Conclusion}
In this paper, we introduce Qwen-Image-VAE-2.0, a suite of high-compression image VAEs ($f16$ and $f32$) designed to overcome the long-standing bottlenecks in native high-resolution image synthesis. Our work demonstrates a clear technical path to resolving the fundamental tripartite trade-off between high compression ratio, reconstruction fidelity, and downstream diffusability: by leveraging expanded latent channel dimensions to compensate for spatial information loss caused by high compression, while simultaneously applying advanced semantic alignment techniques to ensure the latent space remains suitable for diffusion modeling. To achieve state-of-the-art reconstruction fidelity in high-compression regimes, we adopt an improved VAE architecture featuring Global Skip Connections (GSC), establishing a direct path for fine-grained detail recovery. This is bolstered by a comprehensive data strategy where we scale the training corpus to billions of images and utilize a specialized synthetic document-rendering pipeline to ensure the legibility of dense text—a traditional failure point for high-compression models. To address the challenge of diffusability inherent in high-dimensional latent space, we introduce an improved semantic alignment strategy. By aligning latent representations with middle-layer feature of DINOv2, we significantly accelerate the convergence of downstream DiT. Finally, to ensure practical utility, we implement a light-encoder paradigm and an attention-free backbone, maintaining high throughput and minimal encoding overhead even at ultra-high resolutions. Extensive evaluations on public benchmarks and OmniDoc-TokenBench demonstrate that Qwen-Image-VAE-2.0 not only preserves exceptional structural and textual integrity but also facilitates efficient generative modeling. These models provide a robust foundation for the next generation of visual synthesis systems, marking a significant milestone in efficient image generation.

\section{Authors}

\textbf{Contributors:} Zekai Zhang\textsuperscript{*}, Deqing Li\textsuperscript{*}, Kuan Cao\textsuperscript{*}, Yujia Wu\textsuperscript{*}, Chenfei Wu\textsuperscript{\textdagger}, Yu Wu, Liang Peng, Hao Meng, Jiahao Li, Jie Zhang, Kaiyuan Gao, Kun Yan, Lihan Jiang, Ningyuan Tang, Shengming Yin, Tianhe Wu, Xiao Xu, Xiaoyue Chen, Yan Shu, Yanran Zhang, Yilei Chen, Yixian Xu, Yuxiang Chen, Zhendong Wang, Zihao Liu, Zikai Zhou, Yiliang Gu, Yi Wang, Xiaoxiao Xu, Lin Qu

\begingroup
  \renewcommand\thefootnote{}
  \footnotetext{\textsuperscript{*}Equal contribution. 
  \quad \textsuperscript{$\dagger$}Corresponding Author.}
  \addtocounter{footnote}{-1}
\endgroup

\clearpage
\bibliography{colm2024_conference}

@inproceedings{deng2009imagenet,
  title={Imagenet: A large-scale hierarchical image database},
  author={Deng, Jia and Dong, Wei and Socher, Richard and Li, Li-Jia and Li, Kai and Fei-Fei, Li},
  booktitle={2009 IEEE conference on computer vision and pattern recognition},
  pages={248--255},
  year={2009},
  organization={IEEE}
}

@inproceedings{repa-e,
  title={Repa-e: Unlocking vae for end-to-end tuning of latent diffusion transformers},
  author={Leng, Xingjian and Singh, Jaskirat and Hou, Yunzhong and Xing, Zhenchang and Xie, Saining and Zheng, Liang},
  booktitle={Proceedings of the IEEE/CVF International Conference on Computer Vision},
  pages={18262--18272},
  year={2025}
}

@article{Karras2018ASG,
  title={A Style-Based Generator Architecture for Generative Adversarial Networks},
  author={Tero Karras and Samuli Laine and Timo Aila},
  journal={2019 IEEE/CVF Conference on Computer Vision and Pattern Recognition (CVPR)},
  year={2018},
  pages={4396-4405},
  url={https://api.semanticscholar.org/CorpusID:54482423}
}

@article{Wu2025TokBenchEY,
  title={TokBench: Evaluating Your Visual Tokenizer before Visual Generation},
  author={Junfeng Wu and Dongliang Luo and Weizhi Zhao and Zhihao Xie and Yuanhao Wang and Junyi Li and Xudong Xie and Yuliang Liu and Xiang Bai},
  journal={ArXiv},
  year={2025},
  volume={abs/2505.18142},
  url={https://api.semanticscholar.org/CorpusID:278886218}
}

@inproceedings{vavae,
  title={Reconstruction vs. generation: Taming optimization dilemma in latent diffusion models},
  author={Yao, Jingfeng and Yang, Bin and Wang, Xinggang},
  booktitle={Proceedings of the Computer Vision and Pattern Recognition Conference},
  pages={15703--15712},
  year={2025}
}

@article{Ouyang2024OmniDocBenchBD,
  title={OmniDocBench: Benchmarking Diverse PDF Document Parsing with Comprehensive Annotations},
  author={Linke Ouyang and Yuan Qu and Hongbin Zhou and Jiawei Zhu and Rui Zhang and Qunshu Lin and Bin Wang and Zhiyuan Zhao and Man Jiang and Xiaomeng Zhao and Jin Shi and Fan Wu and Pei Chu and Ming-Hao Liu and Zhenxiang Li and Chaoming Xu and Bo Zhang and Botian Shi and Zhongying Tu and Conghui He},
  journal={2025 IEEE/CVF Conference on Computer Vision and Pattern Recognition (CVPR)},
  year={2024},
  pages={24838-24848},
  url={https://api.semanticscholar.org/CorpusID:274609934}
}

@misc{cui2025paddleocr30technicalreport,
      title={PaddleOCR 3.0 Technical Report}, 
      author={Cheng Cui and Ting Sun and Manhui Lin and Tingquan Gao and Yubo Zhang and Jiaxuan Liu and Xueqing Wang and Zelun Zhang and Changda Zhou and Hongen Liu and Yue Zhang and Wenyu Lv and Kui Huang and Yichao Zhang and Jing Zhang and Jun Zhang and Yi Liu and Dianhai Yu and Yanjun Ma},
      year={2025},
      eprint={2507.05595},
      archivePrefix={arXiv},
      primaryClass={cs.CV},
      url={https://arxiv.org/abs/2507.05595}, 
}

@article{cosmos,
  title={Cosmos world foundation model platform for physical ai},
  author={Agarwal, Niket and Ali, Arslan and Bala, Maciej and Balaji, Yogesh and Barker, Erik and Cai, Tiffany and Chattopadhyay, Prithvijit and Chen, Yongxin and Cui, Yin and Ding, Yifan and others},
  journal={arXiv preprint arXiv:2501.03575},
  year={2025}
}

@article{wan2025wan,
  title={Wan: Open and advanced large-scale video generative models},
  author={Wan, Team and Wang, Ang and Ai, Baole and Wen, Bin and Mao, Chaojie and Xie, Chen-Wei and Chen, Di and Yu, Feiwu and Zhao, Haiming and Yang, Jianxiao and others},
  journal={arXiv preprint arXiv:2503.20314},
  year={2025}
}

@article{hunyuanvideo1.5,
  title={Hunyuanvideo 1.5 technical report},
  author={Wu, Bing and Zou, Chang and Li, Changlin and Huang, Duojun and Yang, Fang and Tan, Hao and Peng, Jack and Wu, Jianbing and Xiong, Jiangfeng and Jiang, Jie and others},
  journal={arXiv preprint arXiv:2511.18870},
  year={2025}
}

@article{stepvideo,
  title={Step-video-t2v technical report: The practice, challenges, and future of video foundation model},
  author={Ma, Guoqing and Huang, Haoyang and Yan, Kun and Chen, Liangyu and Duan, Nan and Yin, Shengming and Wan, Changyi and Ming, Ranchen and Song, Xiaoniu and Chen, Xing and others},
  journal={arXiv preprint arXiv:2502.10248},
  year={2025}
}

@article{ltx,
  title={Ltx-video: Realtime video latent diffusion},
  author={HaCohen, Yoav and Chiprut, Nisan and Brazowski, Benny and Shalem, Daniel and Moshe, Dudu and Richardson, Eitan and Levin, Eran and Shiran, Guy and Zabari, Nir and Gordon, Ori and others},
  journal={arXiv preprint arXiv:2501.00103},
  year={2024}
}

@inproceedings{sit,
  title={Sit: Exploring flow and diffusion-based generative models with scalable interpolant transformers},
  author={Ma, Nanye and Goldstein, Mark and Albergo, Michael S and Boffi, Nicholas M and Vanden-Eijnden, Eric and Xie, Saining},
  booktitle={European Conference on Computer Vision},
  pages={23--40},
  year={2024},
  organization={Springer}
}

@article{cfg,
  title={Classifier-free diffusion guidance},
  author={Ho, Jonathan and Salimans, Tim},
  journal={arXiv preprint arXiv:2207.12598},
  year={2022}
}

@inproceedings{dit,
  title={Scalable diffusion models with transformers},
  author={Peebles, William and Xie, Saining},
  booktitle={Proceedings of the IEEE/CVF international conference on computer vision},
  pages={4195--4205},
  year={2023}
}

@misc{flux2024,
    author={Black Forest Labs},
    title={FLUX},
    year={2024},
    howpublished={\url{https://github.com/black-forest-labs/flux}},
}

@misc{qwen-image-2.0,
      title={Qwen-Image-2.0 Technical Report}, 
      author={Bing Zhao and Chenfei Wu and Deqing Li and Hao Meng and Jiahao Li and Jie Zhang and Jingren Zhou and Junyang Lin and Kaiyuan Gao and Kuan Cao and Kun Yan and Liang Peng and Lihan Jiang and Niantong Li and Ningyuan Tang and Shengming Yin and Tianhe Wu and Xiao Xu and Xiaoyue Chen and Xihua Wang and Yan Shu and Yanran Zhang and Yi Wang and Yilei Chen and Ying Ba and Yixian Xu and Yujia Wu and Yuxiang Chen and Zecheng Tang and Zekai Zhang and Zhendong Wang and Zihao Liu and Zikai Zhou and An Yang and Chen Cheng and Chenxu Lv and Dayiheng Liu and Fan Zhou and Hantian Xiong and Hongzhu Shi and Hu Wei and Huihong Zhao and Ivy Liu and Jianwei Zhang and Jiawei Zhang and Kai Chen and Kang He and Levon Xue and Lin Qu and Linhan Tang and Luwen Feng and Minggang Wu and Minmin Sun and Na Ni and Rui Men and Shuai Bai and Sishou Zheng and Tao Lan and Tianqi Zhang and Tingkun Wen and Wei Wang and Weixu Qiao and Weiyi Lu and Wenmeng Zhou and Xiaodong Deng and Xiaoxiao Xu and Xinlei Fang and Xionghui Chen and Yanan Wang and Yang Fan and Yichang Zhang and Yixuan Xu and Yu Wu and Zhiyuan Ma and Zhizhi Cai},
      year={2026},
      eprint={2605.10730},
      archivePrefix={arXiv},
      primaryClass={cs.CV},
      url={https://arxiv.org/abs/2605.10730}, 
}

@inproceedings{sd3,
  title={Scaling rectified flow transformers for high-resolution image synthesis},
  author={Esser, Patrick and Kulal, Sumith and Blattmann, Andreas and Entezari, Rahim and M{\"u}ller, Jonas and Saini, Harry and Levi, Yam and Lorenz, Dominik and Sauer, Axel and Boesel, Frederic and others},
  booktitle={Forty-first international conference on machine learning},
  year={2024}
}

@article{qwenimage,
  title={Qwen-image technical report},
  author={Wu, Chenfei and Li, Jiahao and Zhou, Jingren and Lin, Junyang and Gao, Kaiyuan and Yan, Kun and Yin, Sheng-ming and Bai, Shuai and Xu, Xiao and Chen, Yilei and others},
  journal={arXiv preprint arXiv:2508.02324},
  year={2025}
}

@inproceedings{rombach2022high,
  title={High-resolution image synthesis with latent diffusion models},
  author={Rombach, Robin and Blattmann, Andreas and Lorenz, Dominik and Esser, Patrick and Ommer, Bj{\"o}rn},
  booktitle={Proceedings of the IEEE/CVF conference on computer vision and pattern recognition},
  pages={10684--10695},
  year={2022}
}

@article{seedream3,
  title={Seedream 3.0 technical report},
  author={Gao, Yu and Gong, Lixue and Guo, Qiushan and Hou, Xiaoxia and Lai, Zhichao and Li, Fanshi and Li, Liang and Lian, Xiaochen and Liao, Chao and Liu, Liyang and others},
  journal={arXiv preprint arXiv:2504.11346},
  year={2025}
}

@inproceedings{dcae,
  title={Deep Compression Autoencoder for Efficient High-Resolution Diffusion Models},
  author={Chen, Junyu and Cai, Han and Chen, Junsong and Xie, Enze and Yang, Shang and Tang, Haotian and Li, Muyang and Han, Song},
  booktitle={The Thirteenth International Conference on Learning Representations},
  year={2024}
}

@inproceedings{diffusability,
  title={Improving the Diffusability of Autoencoders},
  author={Skorokhodov, Ivan and Girish, Sharath and Hu, Benran and Menapace, Willi and Li, Yanyu and Abdal, Rameen and Tulyakov, Sergey and Siarohin, Aliaksandr},
  booktitle={Forty-second International Conference on Machine Learning},
  year={2025}
}

@article{dinov2,
  title={Dinov2: Learning robust visual features without supervision},
  author={Oquab, Maxime and Darcet, Timoth{\'e}e and Moutakanni, Th{\'e}o and Vo, Huy and Szafraniec, Marc and Khalidov, Vasil and Fernandez, Pierre and Haziza, Daniel and Massa, Francisco and El-Nouby, Alaaeldin and others},
  journal={arXiv preprint arXiv:2304.07193},
  year={2023}
}

@inproceedings{dcae1.5,
  title={Dc-ae 1.5: Accelerating diffusion model convergence with structured latent space},
  author={Chen, Junyu and Zou, Dongyun and He, Wenkun and Chen, Junsong and Xie, Enze and Han, Song and Cai, Han},
  booktitle={Proceedings of the IEEE/CVF International Conference on Computer Vision},
  pages={19628--19637},
  year={2025}
}

@article{attention,
  title={Attention is all you need},
  author={Vaswani, Ashish and Shazeer, Noam and Parmar, Niki and Uszkoreit, Jakob and Jones, Llion and Gomez, Aidan N and Kaiser, {\L}ukasz and Polosukhin, Illia},
  journal={Advances in neural information processing systems},
  volume={30},
  year={2017}
}

@article{dinov3,
  title={Dinov3},
  author={Sim{\'e}oni, Oriane and Vo, Huy V and Seitzer, Maximilian and Baldassarre, Federico and Oquab, Maxime and Jose, Cijo and Khalidov, Vasil and Szafraniec, Marc and Yi, Seungeun and Ramamonjisoa, Micha{\"e}l and others},
  journal={arXiv preprint arXiv:2508.10104},
  year={2025}
}

@article{bolya2025perception,
  title={Perception encoder: The best visual embeddings are not at the output of the network},
  author={Bolya, Daniel and Huang, Po-Yao and Sun, Peize and Cho, Jang Hyun and Madotto, Andrea and Wei, Chen and Ma, Tengyu and Zhi, Jiale and Rajasegaran, Jathushan and Rasheed, Hanoona and others},
  journal={arXiv preprint arXiv:2504.13181},
  year={2025}
}

@inproceedings{mae,
  title={Masked autoencoders are scalable vision learners},
  author={He, Kaiming and Chen, Xinlei and Xie, Saining and Li, Yanghao and Doll{\'a}r, Piotr and Girshick, Ross},
  booktitle={Proceedings of the IEEE/CVF conference on computer vision and pattern recognition},
  pages={16000--16009},
  year={2022}
}

@inproceedings{lpips,
  title={The unreasonable effectiveness of deep features as a perceptual metric},
  author={Zhang, Richard and Isola, Phillip and Efros, Alexei A and Shechtman, Eli and Wang, Oliver},
  booktitle={Proceedings of the IEEE conference on computer vision and pattern recognition},
  pages={586--595},
  year={2018}
}

@article{vtp,
  title={Towards Scalable Pre-training of Visual Tokenizers for Generation},
  author={Yao, Jingfeng and Song, Yuda and Zhou, Yucong and Wang, Xinggang},
  journal={arXiv preprint arXiv:2512.13687},
  year={2025}
}

@article{rae,
  title={Diffusion transformers with representation autoencoders},
  author={Zheng, Boyang and Ma, Nanye and Tong, Shengbang and Xie, Saining},
  journal={arXiv preprint arXiv:2510.11690},
  year={2025}
}

@article{ltx2,
  title={LTX-2: Efficient Joint Audio-Visual Foundation Model},
  author={HaCohen, Yoav and Brazowski, Benny and Chiprut, Nisan and Bitterman, Yaki and Kvochko, Andrew and Berkowitz, Avishai and Shalem, Daniel and Lifschitz, Daphna and Moshe, Dudu and Porat, Eitan and others},
  journal={arXiv preprint arXiv:2601.03233},
  year={2026}
}

@article{hunyuanvideo,
  title={Hunyuanvideo: A systematic framework for large video generative models},
  author={Kong, Weijie and Tian, Qi and Zhang, Zijian and Min, Rox and Dai, Zuozhuo and Zhou, Jin and Xiong, Jiangfeng and Li, Xin and Wu, Bo and Zhang, Jianwei and others},
  journal={arXiv preprint arXiv:2412.03603},
  year={2024}
}

@article{hunyuanimage3.0,
  title={Hunyuanimage 3.0 technical report},
  author={Cao, Siyu and Chen, Hangting and Chen, Peng and Cheng, Yiji and Cui, Yutao and Deng, Xinchi and Dong, Ying and Gong, Kipper and Gu, Tianpeng and Gu, Xiusen and others},
  journal={arXiv preprint arXiv:2509.23951},
  year={2025}
}

@misc{hunyuanimage2.1,
  title={HunyuanImage 2.1: An Efficient Diffusion Model for High-Resolution (2K) Text-to-Image Generation},
  author={Tencent Hunyuan Team},
  year={2025},
  howpublished={\url{https://github.com/Tencent-Hunyuan/HunyuanImage-2.1}},
}

@misc{flux2,
    author={Black Forest Labs},
    title={{FLUX.2: Frontier Visual Intelligence}},
    year={2025},
    howpublished={\url{https://bfl.ai/blog/flux-2}},
}

@article{qiu2025image,
  title={Image tokenizer needs post-training},
  author={Qiu, Kai and Li, Xiang and Chen, Hao and Kuen, Jason and Xu, Xiaohao and Gu, Jiuxiang and Luo, Yinyi and Raj, Bhiksha and Lin, Zhe and Savvides, Marios},
  journal={arXiv preprint arXiv:2509.12474},
  year={2025}
}

@inproceedings{gan,
  title={Image-to-image translation with conditional adversarial networks},
  author={Isola, Phillip and Zhu, Jun-Yan and Zhou, Tinghui and Efros, Alexei A},
  booktitle={Proceedings of the IEEE conference on computer vision and pattern recognition},
  pages={1125--1134},
  year={2017}
}

@article{ssim,
  title={Image quality assessment: from error visibility to structural similarity},
  author={Wang, Zhou and Bovik, Alan C and Sheikh, Hamid R and Simoncelli, Eero P},
  journal={IEEE transactions on image processing},
  volume={13},
  number={4},
  pages={600--612},
  year={2004},
  publisher={IEEE}
}

@inproceedings{psnr,
  title={Image quality metrics: PSNR vs. SSIM},
  author={Hore, Alain and Ziou, Djemel},
  booktitle={2010 20th international conference on pattern recognition},
  pages={2366--2369},
  year={2010},
  organization={IEEE}
}

@article{fid,
  title={Gans trained by a two time-scale update rule converge to a local nash equilibrium},
  author={Heusel, Martin and Ramsauer, Hubert and Unterthiner, Thomas and Nessler, Bernhard and Hochreiter, Sepp},
  journal={Advances in neural information processing systems},
  volume={30},
  year={2017}
}

@article{inception,
  title={Improved techniques for training gans},
  author={Salimans, Tim and Goodfellow, Ian and Zaremba, Wojciech and Cheung, Vicki and Radford, Alec and Chen, Xi},
  journal={Advances in neural information processing systems},
  volume={29},
  year={2016}
}

@article{Liu2019ICDAR2R,
  title={ICDAR 2019 Robust Reading Challenge on Reading Chinese Text on Signboard},
  author={Xi Liu and Rui Zhang and Yongsheng Zhou and Qianyi Jiang and Qi Song and Nan Li and Kai Zhou and Lei Wang and Dong Wang and Minghui Liao and Mingkun Yang and Xiang Bai and Baoguang Shi and Dimosthenis Karatzas and Shijian Lu and C. V. Jawahar},
  journal={2019 International Conference on Document Analysis and Recognition (ICDAR)},
  year={2019},
  pages={1577-1581},
  url={https://api.semanticscholar.org/CorpusID:209439793}
}

@article{Marzal1993ComputationON,
  title={Computation of Normalized Edit Distance and Applications},
  author={Andr{\'e}s Marzal and Enrique Vidal},
  journal={IEEE Trans. Pattern Anal. Mach. Intell.},
  year={1993},
  volume={15},
  pages={926-932},
  url={https://api.semanticscholar.org/CorpusID:14851115}
}
\bibliographystyle{colm2024_conference}

\end{document}